%% file: main.tex
\documentclass[10pt,twocolumn,letterpaper]{article}

\usepackage{cvpr}              % To produce the CAMERA-READY version
\input{preamble}

\definecolor{cvprblue}{rgb}{0.21,0.49,0.74}
\usepackage[pagebackref,breaklinks,colorlinks,citecolor=cvprblue]{hyperref}

\usepackage{graphicx}
\usepackage{enumitem}
\usepackage[utf8]{inputenc} % allow utf-8 input
\usepackage[T1]{fontenc}    % use 8-bit T1 fonts
\usepackage{url}            % simple URL typesetting
\usepackage{booktabs}       % professional-quality tables
\usepackage{amsfonts}       % blackboard math symbols
\usepackage{nicefrac}       % compact symbols for 1/2, etc.
\usepackage{microtype}      % microtypography
\usepackage{fontawesome5}
\usepackage{amsmath}
\usepackage{bm}
\usepackage{bbm}
\usepackage{amssymb}
\usepackage{multirow}

\makeatletter
\newcommand*{\overrightharpoonup}{\mathpalette{\overarrow@\rightharpoonupfill@}}
\newcommand*{\rightharpoonupfill@}{\arrowfill@\relbar\relbar\rightharpoonup}
\makeatother
\DeclareFontFamily{U}{matha}{\hyphenchar\font45}
\DeclareFontShape{U}{matha}{m}{n}{
      <5> <6> <7> <8> <9> <10> gen * matha
      <10.95> matha10 <12> <14.4> <17.28> <20.74> <24.88> matha12
      }{}
\DeclareSymbolFont{matha}{U}{matha}{m}{n}
\DeclareMathSymbol{\varrightharpoonup}{3}{matha}{"E1}

\newcommand{\normv}[1]{\raisebox{0.4ex}{$\rVert$}{#1}\raisebox{0.4ex}{$\lVert$}}

\def\paperID{8732} % *** Enter the Paper ID here
\def\confName{CVPR}
\def\confYear{2024}

\title{Traffic Video Object Detection using Motion Prior}

\author{Lihao Liu$^{1}$,\: Yanqi Cheng$^{1}$,\: Dongdong Chen$^{2}$,\: Jing He$^{3}$,\: \\ Pietro Liò$^{1}$,\: Carola-Bibiane Schönlieb$^{1}$, Angelica I Aviles-Rivero$^{1}$
\\  \:
$^{1}$ University of Cambridge \,\,
$^{2}$ Microsoft Cloud + AI \,\,
$^{3}$ GSK \,\,\,\,
}

\begin{document}

\twocolumn[{
\maketitle
\begin{center}
    \vspace{-2mm}
    \captionsetup{type=figure}
    \includegraphics[width=1\linewidth]{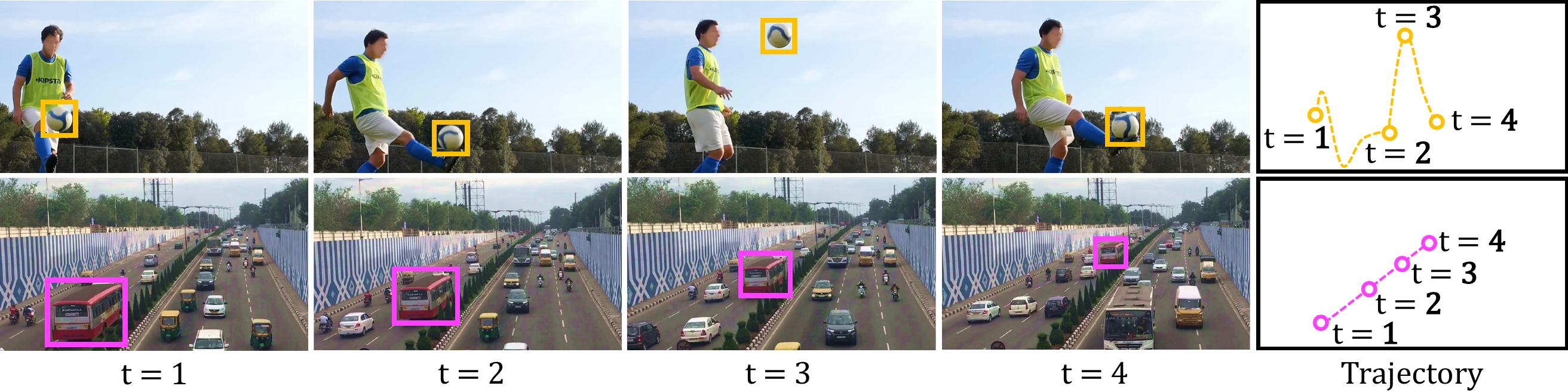}
    \vspace{-6mm}
    \captionof{figure}{Comparative representation of motion trajectories in general scenes (first row) and traffic scenes (second row). In the general scenes, the motion trajectories are irregular and lack a predictable pattern due to the erratic movement of a soccer ball. In contrast, traffic scenes display a more predictable and structured movement pattern. Here, vehicles typically move in a straight, along-the-road direction.}
    \label{fig:motion_prior}
\end{center}
}]

\input{section/abstract}

\input{section/introduction}

\input{section/related_work}

\input{section/method}

\input{section/experiment}

\input{section/conclusion}

{
    \small
    \bibliographystyle{ieeenat_fullname}
    \bibliography{main}
}


\end{document}

%% file: preamble.tex
\usepackage[dvipsnames]{xcolor}

%% file: section/abstract.tex
\begin{abstract}

Traffic videos inherently differ from generic videos in their stationary camera setup, thus providing a strong motion prior where objects often move in a specific direction over a short time interval. Existing works predominantly employ generic video object detection framework for traffic video object detection, which yield certain advantages such as broad applicability and robustness to diverse scenarios. However, they fail to harness the strength of motion prior to enhance detection accuracy. In this work, we propose two innovative methods to exploit the motion prior and boost the performance of both fully-supervised and semi-supervised traffic video object detection. Firstly, we introduce a new self-attention module that leverages the motion prior to guide temporal information integration in the fully-supervised setting. Secondly, we utilise the motion prior to develop a pseudo-labelling mechanism to eliminate noisy pseudo labels for the semi-supervised setting. Both of our motion-prior-centred methods consistently demonstrates superior performance, outperforming existing state-of-the-art approaches by a margin of 2\% in terms of mAP.

\end{abstract}

%% file: section/introduction.tex
\section{Introduction} \label{sec:introduction}

Video object detection~\cite{jiao2021new,zhu2020review,zhou2022transvod,wu2019sequence} is a challenging and fast-progressing computer vision task. Different from image object detection, it incorporates temporal information across frames~\cite{zhu2017deep,zhu2017flow,zhu2018towards,liu2022scotch,cheng2022deep} to improve object detection accuracy. It has been widely applied in diverse fields such as autonomous vehicles, sports analysis, and human-computer interaction~\cite{ghahremannezhad2023object}. 
Hence, this leads to significant interest in developing video object detection methods, which the current body of literature reports outstanding performance.

Traffic video object detection~\cite{ghahremannezhad2023object,ghahremannezhad2022traffic,boukerche2021object,chandrakar2022enhanced}, a specialised use of video object detection, plays a significant role in traffic monitoring and management. Compared to generic videos, traffic videos are often captured by fixed-position cameras installed on roads. As a result, traffic videos often exhibit a \textbf{\textit{motion prior}}, indicating the predictable movement of objects in a specific direction over short time intervals; as shown in Figure~\ref{fig:motion_prior}. By applying video object detection techniques to these traffic videos, various objects of interest in traffic scenes can be identified and tracked, such as vehicles, pedestrians, and road signs. Furthermore, the valuable insights gained from traffic video object detection can help improve traffic management, safety, and overall efficiency in both urban and highway settings. In this work, our main focus is traffic video object detection. 

Most current traffic video object detection methods~\cite{abdulrahim2016cumulative,rauf2016pedestrian,zhu2017deep,zhu2017flow,wu2019sequence,gong2021temporal,garcia2020background,shi2022unsupervised} directly adapt existing deep learning models designed for general video scenarios, such as Faster R-CNN~\cite{ren2015faster} and YOLO~\cite{redmon2016you}. Despite commendable performance achieved, these techniques are not specifically tailored for traffic scenarios, which could potentially limit their effectiveness in fully leveraging the unique motion prior of traffic videos.

Motivated by these considerations, in this work, we argue that adding motion prior extracted from traffic scenes can substantially enhance the performance of traffic video object detection. To this end, we delve into the realm of embedding motion prior within both fully-supervised and semi-supervised contexts. Specifically, in the fully-supervised setting, we first develop a self-attention module that overlays a motion prior mask on attention maps. This new self-attention module advances performance by fostering inter-frame temporal information integration. In addition, in the semi-supervised setting, we introduce a pseudo label filtering strategy that rectifies imprecise pseudo labels through a motion prior filter, further enhancing the quality of pseudo labels. Our contributions are summarized as follows:
\begin{itemize}
\item We propose embedding motion prior from traffic scenes into the design of our object detection models in both fully-supervised and semi-supervised settings. This allows the models to better interpret the unique motion prior of traffic videos and improves object detection performance.
    
    \begin{itemize}
    \item We design a novel self-attention module that applies a motion prior mask on attention maps, that helps establish temporal information integration between frames.
    
    \item We introduce a pseudo label filtering strategy that employs a motion prior filter to refine inaccurate pseudo labels in the semi-supervised setting. 
    \end{itemize}

\item We evaluate the proposed two methods on the traffic benchmark dataset TrafficCAM~\cite{deng2022trafficcam}, and compare with the state-of-the-art methods. Quantitative and qualitative experimental results show that our methods outperform existing ones on traffic video object detection by a margin of 2\% mAP in both fully- and semi-supervised settings. 
\end{itemize}

%% file: section/related_work.tex
\section{Related Work}

\hspace*{4mm}\textbf{Image Object Detection in Traffic Scenario.}
With the advent of deep learning, CNN-based techniques have been well developed in the field of object detection which are widely applied in traffic-related applications. There are several milestones for image object detection e.g. YOLO~\cite{redmon2016you,asha2018vehicle}, and SSD~\cite{liu2016ssd,wang2018multi}, which directly predicting object class and bounding box regression
in a single stage.

Despite the one stage methods, R-CNN family~\cite{girshick2015fast,ren2015faster,othmani2022vehicle} are proposed with two staged detection, which introduces the Region Proposal Network (RPN) to propose Region of Interest (RoI) for guiding the detection. Faster RCNN~\cite{ren2015faster} is one of the most popular image-level algorithms that achieve outstanding results. Based on it, Libra~\cite{pang2019libra}, Guided Anchoring~\cite{wang2019region}, Dynamic R-CNN~\cite{zhang2020dynamic}, and SABL~\cite{wang2020side} extend anchor mechanism to improve the accuracy of object detection by better handling the object scale variation and class imbalance. Double Heads~\cite{wu2020rethinking} implements two parallel  detection heads to allow further fine-grained control over the detection process. Although these methods already gain strong performance, they have not considered the inter-frame connection in traffic videos.

\smallskip
\textbf{Traffic Video Object Detection.} 
Traffic video analysis has gained increasing interest in recent years~\cite{ghahremannezhad2023object}, an intuitive research line is to use above image object detection models to process traffic videos frame-by-frame. Another body of works have investigated the integration of temporal information, which can be divided into three categories: frame differencing~\cite{abdulrahim2016cumulative}, feature integration~\cite{rauf2016pedestrian,zhu2017deep,zhu2017flow,wu2019sequence,gong2021temporal}, and background subtraction modelling~\cite{garcia2020background,shi2022unsupervised}. Among the three categories, feature integration methods have emerged as a prominent technique.

\smallskip
\textbf{Feature Integration for Video Analysis.}
%\subsection{Feature Integration for Video Analysis}
A common strategy for feature integration is to integrate features from other video frames to the current frame. In the work of DFF~\cite{zhu2017deep} and FGFA~\cite{zhu2017flow}, optical flow~\cite{gibson2016optical} is applied, which calculates the pixel vector shift between consecutive frames. SELSA~\cite{wu2019sequence} and TRA~\cite{gong2021temporal} employ a self-attention mechanism~\cite{vaswani2017attention} that addresses temporal relation between RoIs across frames. Although these methods incorporate video-based mechanisms in the network, they are not traffic-specific, making it challenging to handle complex traffic conditions, e.g. motion blur, lighting changes, noisy backgrounds. 
Hence, in our work, we will explore how to embed the inherent motion prior from traffic videos into a self-attention module, aiming to enhance feature integration.

\begin{figure*}[ht]
    \centering
    \includegraphics[width=0.88\linewidth]{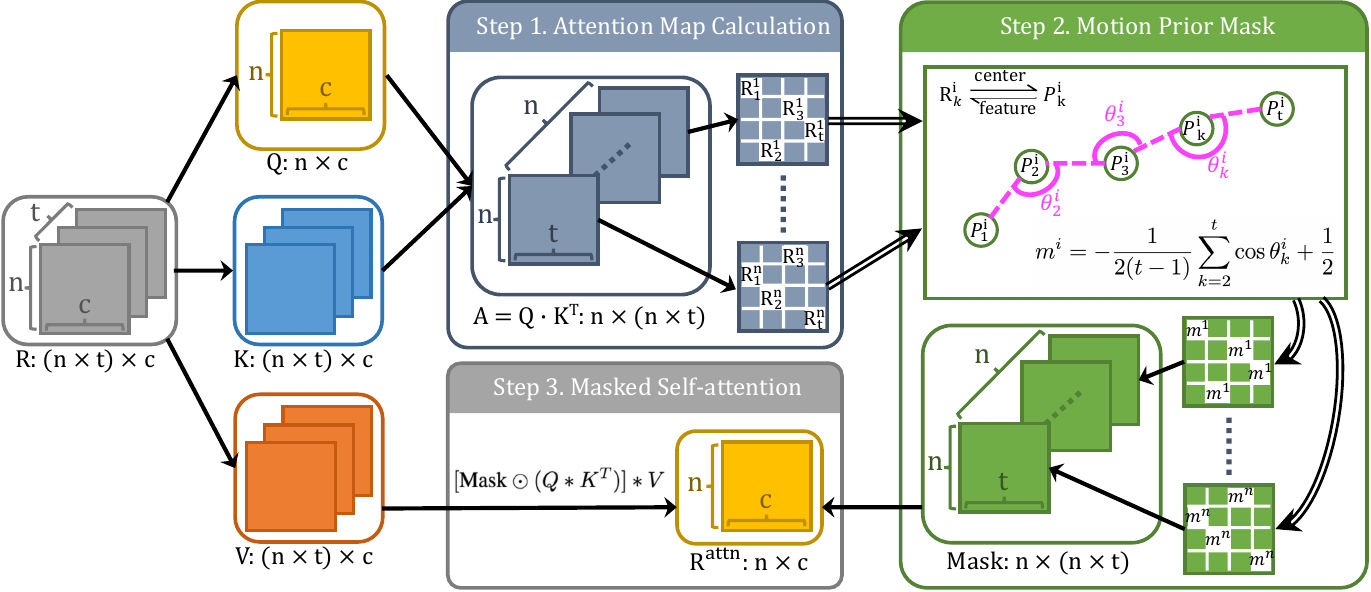}
    \vspace{-1mm}
    \caption{
    Motion Prior Attention. The video-level RoI features $R$ serve as the basis for generating $Q, K, V$, where $Q$ derives from the first frame RoI features, while $K, V$ are obtained from the complete RoI features. In Step 1, $Q$ and $K$ are used to compute the similarity metrics $A$. For each RoI in the initial frames, the mechanism identifies the most similar RoI feature in each of the $t$ frames in the video. These selected RoI feature maps with high similarity are then forwarded to Step 2 for individual processing. In Step 2, the motion prior is employed to verify if the respective centres of the RoIs align along a straight trajectory. A score, $m^i$, is derived to evaluate the likelihood of the RoIs aligning in a straight line. This results in an $A$-shaped mask that accentuates the RoIs aligned along a straight trajectory. In Step 3, the Mask, $A$, and $V$ collaborate to integrate the video information into a single frame, encapsulating the essence of the entire sequence into $R^{attn}$. } 
    \label{fig:self_attention_module}  
    \vspace{-3mm}
\end{figure*}

\smallskip
\textbf{Pseudo-Labelling Mechanisms for Semi-Supervised Video Object Detection.}
A common limitation of the above methods is their dependence on substantial labeled data, which is costly and time-consuming to obtain, especially given complex traffic scenarios like occlusion and object diversity. This has prompted the exploration of alternative strategies, notably semi-supervised video object detection~\cite{yan2019semi}, that enhance detection performance while lessening the requirement for extensive labeled data. The most popular works in semi-supervised area can be categorised into two groups: consistency-based learning~\cite{jeong2019consistency} and pseudo label learning~\cite{sohn2020simple,ke2022three}. Consistency-based learning methods focus on encouraging the model to produce consistent predictions for different augmentations of the same unlabelled input~\cite{tarvainen2017mean,xie2020unsupervised,sohn2020fixmatch}. This strategy allows the model to exploit the underlying structure of the data and improve its generalisation capabilities. On the other hand, a more effective strategy in object detection area is called pseudo-labelling~\cite{xu2021end,sohn2020simple}. Pseudo-labelling approaches involve a two-stage process. In the first stage, the model trains on labelled data and generates pseudo labels for the unlabelled data. Then, in the second stage, the model is re-trained using both labelled data and pseudo labels for further refinement. In our work, we also explore how to effectively use the motion prior found in traffic videos to improve object detection performance in a semi-supervised setting.

%% file: section/method.tex
\section{Methodology} \label{sec:method}
In this section, we present two key elements: a motion-based self-attention module and a pseudo label filtering strategy.

\subsection{Motion Prior Attention} 

\hspace*{4mm}\textbf{Object Detection Workflow.} 
We adopt SELSA~\cite{wu2019sequence} as our baseline for traffic video object detection, which comprises three stages: (i) A region proposal network (RPN)~\cite{ren2015faster} that accepts a video clip as input, and outputs multiple features in every frame. Each feature indicates a potential region that may contain target objects, referred as RoI features; (ii) A feature integration module that constructs temporal relationships for the RoI features to enhance the overall performance; and (iii) A bounding box regression and object classification branch, which process the integrated feature to generate the final bounding boxes and class labels. In our work, we focus on stage (ii) and developing a novel self-attention module that optimises the integration of features using motion priors.

\textbf{Self-Attention and Feature Integration}. Self-attention is a powerful mechanism used to capture the relationships between different elements in a sequence. A typical self-attention computes the query, key, and value matrices ($Q$, $K$, $V$) from the input data. Then, it calculates attention scores by taking the dot product of query and key matrices, which are then used to compute weighted values. For feature integration purposes, $Q$ is often taken as a sub-sequence of the $K$. This arrangement allows the self-attention module to capture the relationships between a specific portion and the entire input. As a result, it enables the integration of all information into a specific portion. 

\faHandPointRight[regular] \textbf{Step 1. Attention Map Calculation.} As shown in the left part of Figure~\ref{fig:self_attention_module}, we start with a Regions of Interest feature (RoI feature) obtained from a Region Proposal Network (RPN). We denote the proposed RoI features as $R \in \mathbb{R}^{(n \times t) \times c}$, where $n$ is the number of RoI features in a single frame, $t$ is the total number of frames, and $c$ is the number of feature channels. To integrate all information into a single specific frame, we extract the RoI features in the current frame $\hat{R} \in \mathbb{R}^{n\times c}$. We then set $Q, K, V$ as $\hat{R}, {R}, {R}$, respectively. Subsequently, we compute the self-attention map using $A = Q * K^T \in \mathbb{R}^{n\times (n\times t)}$.

\faHandPointRight[regular] \textbf{Step 2. Motion Prior Mask.} 
The attention map $A$ serves as a similarity metric between $\hat{R}$ and ${R}$. Hence, for each RoI feature in the current frame $\hat{R}^i$, $A$ can be used to identify the RoI feature most similar to $\hat{R}^i$ in each of the $t$ frames, which we denote as ${R}^{i}_{jk}$, where $j \in \{1,2,3,...,n\}$ and $k \in \{1,2,3,...,t\}$. The index of the most similar RoI features to $\hat{R}^i$ to in the $k$ frame can be computed by
${o}_k = \operatorname*{argmax}_{j} A^{i}_{jk}$.
Consequently, for every single RoI feature in the current frame, we obtain a list of RoI features $[R_{{o_1}1}^i, R_{{o_2}2}^i, R_{{o_3}3}^i, ..., R_{{o_t}t}^i]$ that form the potential moving trajectory of an object across the video. For simplicity, we omit $o_1, o_2, o_3, ..., o_t$ in our notation, since $o_k$ varies with $k$, and maintain the list as $[R_{1}^i, R_{2}^i, R_{3}^i, ..., R_{t}^i]$.
After obtaining the RoI feature list, we get the exact centres for the RoI $[P_1^i, P_2^i, ..., P_t^i]$, each point represented by its coordinates, i.e. $P_k^i = (x_k^i, y_k^i)$; see Step 2 in Figure~\ref{fig:self_attention_module}.%

Denote the angle between $\overrightharpoonup{P_{k}^iP_{k-1}^i}$ and $\overrightharpoonup{P_{k}^iP_{k+1}^i}$ as $\theta_k^i$. We calculate the cosine value of this angle to determine whether the centres of RoI aligned in a straight line through:
\begin{equation}
    \cos{\theta_{k}^i}= \frac{\overrightharpoonup{P_{k}^iP_{k-1}^i} \cdot \overrightharpoonup{P_{k}^iP_{k+1}^i}}{\normv{\overrightharpoonup{P_{k}^iP_{k-1}^i}} \, \times \,\normv{\overrightharpoonup{P_{k}^iP_{k+1}^i}}} 
\end{equation}
where $\,\cdot\,$ is the vector dot product, $\normv{\overrightharpoonup{P_{k}^iP_{k-1}^i}}$ and $\normv{\overrightharpoonup{P_{k}^iP_{k+1}^i}}$ is the magnitudes of the vectors $\overrightharpoonup{P_{k}^iP_{k-1}^i}$ and $\overrightharpoonup{P_{k}^iP_{k+1}^i}$, respectively.
If $\cos{\theta_k^i}= -1 \, (0 \leqslant \theta_k^i < 2\pi)$, then $\theta_k^i= \pi$, \textit{i.e.} the angle between vectors $\overrightharpoonup{P_{k}^iP_{k-1}^i}$ and $\overrightharpoonup{P_{k}^iP_{k+1}^i}$ is $180^{\circ}$, thus $P_{k-1}^i,\, P_{k}^i,\, P_{k+1}^i$ align in a straight line.

To quantify the probability of the moving trajectory being in a straight line, we sum up the cosine scores of $\theta_k^i \, \, ,\forall \, k \in \{2,3,...,t\}$,
and normalise the $\cos \theta_k^i$ to the range $[0, 1]$:
\begin{equation} \label{eqn:score}
   m^i = - \frac{1}{2(t-1)} \sum^{t}_{k=2} \cos{\theta_{k}^i} + \frac{1}{2} 
\end{equation}
where a high alignment $m^i$ close to 1 means the points are well-aligned, and vice versa.%
We then use this $m^i$ to generate a mask for the self-attention map $A$. This attention map mask, denoted as "$\text{Mask}$", can be obtained as follows:
\begin{equation}
\text{Mask}_{jk}^i= \left\{
    \begin{aligned}
        &\,\,m^i, \, \, \, \forall \, k \in \{1,2,\ldots,t\} \,\, \text{and} \,\, j=o_k \\
        &\,\,1- m^i, \quad\quad\quad\quad\quad\quad\quad \text{otherwise} \,
    \end{aligned} 
\right.
\end{equation}
where "$\text{Mask}$" shares the same dimensions as the self-attention map $A$. If the high-similarity RoIs are likely to align along a straight line, then the mask will enhance these RoIs in the self-attention map $A$. Otherwise, the mask will diminish their prominence in the self-attention map.

\faHandPointRight[regular] \textbf{Step 3. Masked Self-Attention.}
By masking this alignment score matrix "$\text{Mask}$" for all RoI features in the attention mask $A$, we can obtain a masked attention map that highlights the RoI features that might be aligned in a straight line. Lastly, we get the motion prior integrated self-attention results by applying the masked attention map back to the $V$: 
\begin{equation}
   {R}^{\text{attn}} = [\text{Mask} \odot (Q * K^T)] * V
\end{equation}
where ${R}^{\text{attn}}$ is the output of the new self-attention module, which is used in the final prediction of the detection model in stage (iii).

\textbf{Cost Function Driven Optimisation.} For the bounding box localization, we use the Huber loss to predict the bounding box. The bounding box is represented as $(x, y, w, h)$, and a detection loss per frame is used in conjunction:
\begin{equation}
\mathcal{L}_{bbox} = \sum_{i=1}^{n} \sum_{e \in \{x, y, w, h\}} \mathcal{L}_{smooth}(\hat{B}^{i}{|}_{e} - B^{i}{|}_{e}),
\end{equation}
\begin{equation}
\,\,  \mathcal{L}_{smooth}(\delta) =
\begin{cases}
0.5 \cdot {\delta}^2 & \text{if } |{\delta}| < 1 \\ 
|{\delta}| - 0.5 & \text{otherwise}
\end{cases}
\end{equation}
where $\hat{B}^{i}{|}_{e}$ and $B^{i}{|}_{e}$ represent the ground truth and predicted values for element $e$ of the $i$-th bounding box, respectively, and $n$ is the number of bounding boxes. By minimizing this loss, the model learns to accurately predict the location of the bounding boxes.

For the classification label prediction of the bounding box, we use Cross Entropy loss to measure the difference between the predicted and ground truth class probabilities. In total, there are 10 classes. Hence, the loss is: 
\begin{equation}
\mathcal{L}_{cls} = -\sum_{i=1}^{n} \sum_{c\in\{1,2,...,10\}} y^{i}{|}_{c} \log(p^{i}{|}_{c})
\end{equation}
where $y^{i}{|}_{c}$ is the binary ground truth label for class $c$ of the $i$-th bounding box, and $p^{i}{|}_{c}$ is the predicted probability for class $c$ of the $i$-th bounding box.

By minimising $\mathcal{L}_{total}$, the model learns to accurately predict the location and category of objects in images:
\begin{equation}
   \mathcal{L}_{total} = \mathcal{L}_{bbox} + \mathcal{L}_{cls}
\end{equation}

\subsection{High-Confidence Learning with Motion-Prior-Enhanced Pseudo-Labeling} \label{subsec:semi-sup}

\hspace{4mm}\textbf{Pseudo-Labelling Mechanism.} 
We select STAC as our pseudo-labelling mechanism, which contains three stages: i) a fully-supervised training stage, that train on labelled dataset, ii) the generation of pseudo labels for the unlabelled data using the saved weights obtained from previous stage, and iii) re-training the model on both the labelled data and the unlabelled data with pseudo labels. In this work, we focused on the stage ii) to improve the quality of generated pseudo labels.

\textbf{Vanishing Point Centred Area.}
The edges of a road are parallel in a 3D world, but converge to a vanishing point in a 2D image taken from a traffic camera~\cite{kong2009vanishing}. Theoretically, trajectories of moving objects on this road should intersect with the road edges at the vanishing point. However, due to slight deviations and other variables, the moving trajectory might not perfect intersect at the vanishing point, but intersect within an area centred around the vanishing point. Therefore, by utilizing this geometrically-inspired motion prior, we propose a four-step pseudo-label filtering strategy to filter out noisy pseudo labels and refine them further. We now explain each of these four steps in detail.

\faHandPointRight[regular] \textbf{Step 1. Generating Pseudo Labels.} We remind the reader that our experimental setup concerns video object detection. In this context, the probability output of a network corresponds  to the probability of a given class being present in a sample. A standard approach then to generate pseudo labels is to apply a threshold to these probabilities values. Formally, let $Y = \{(B^i, C^i, S^i)\}_{i=1}^{N^*}$ be the set of pseudo labels (RoIs) for the unlabelled data, where $B^i$ denotes the bounding box, $C^i$ the predicted class, and $S^i$ the confidence score of the $i$-th pseudo label. A common confidence  approach to remove labels with low confidence scores is expressed as $\hat{Y} = \{(B^i, C^i, S^i) \in Y | S^i \geq \sigma\}$, where $\sigma\in [0,1]$ is a threshold that yields to hard pseudo labels. However, these approaches do not modify the predicted class accordingly. To address this issue and refine the noisy pseudo labels, we use above mentioned motion prior to filter the pseudo labels and refine them further. 

\faHandPointRight[regular] \textbf{Step 2. Producing Trajectories from Pseudo Labels.}
For each frame $F_k$ of the unlabeled video, let $Y_k = \{(B^{i}_{k}, C^{i}_{k}, S^{i}_{k})\}_{i=1}^{n_k}$ be the set of RoIs, where ${n_k}$ is the RoI numbers in $F_k$. For each $B^{i}_{k}$, we find the bounding box $B^{i'}_{k+1}$ in the next frame $F_{k+1}$ that maximises the overlap, i.e., $i' = \operatorname*{argmax}_{\hat{i}} \text{IoU}(B^{i}_{k}, B^{\hat{i}}_{k+1})$, for $\hat{i} \in \{1,2,...,n_{k+1}\}$, where $\text{IoU}(\cdot)$ denotes the intersection over union. By repeating this process within a short video clip starting from $F_k$ of length $d$, we obtain a set of moving trajectories $\bm{H_k} = \{H_i\}_{i=1}^{n_k}$, where $H_i = \{B^i_k, \, B^{i'}_{k{+}1}, \, B^{i''}_{k{+}2}, ..., \, B^{{i''}^{..}{''}}_{k{+}d}\}$.
This then leads to trajectories in the video with $\bm{H} = \cup_{k} \bm{H_k}$ where $\forall k \in \{d, 2d, 3d, ... ,t{-}d \}$, which after re-indexing gives $\bm{H} = \{H_i\}_{i=1}^{N}$, where $N$ is the trajectory number in the entire video. Normally $N < N^*$, since the number of trajectories is smaller than the number of RoIs.

\faHandPointRight[regular] \textbf{Step 3. Identify Vanishing Point Centred Area.}
For each moving trajectory $H_i$, we use linear regression to obtain a straight line approximation $L_i$ given by the equation $y = \beta_i x + \alpha_i$, where $\beta_i$ is the slope of the line and $\alpha_i$ is its intercept. The intersection points $P_{i{\bar{i}}}$ between two non-parallel trajectories $L_i$ and $L_{\bar{i}}$ can be obtained by solving the system of equations given by the intersection of their respective lines $L_i$ and $L_{\bar{i}}$, i.e., $y = \beta_i x + \alpha_i$ and $y = \beta_{\bar{i}} x + \alpha_{\bar{i}}$. We apply the DBSCAN clustering algorithm~\cite{ester1996density} on the set of intersection points $\bm{P}$ to identify an area that densely aggregates intersection points (vanishing point centred area), denoted as $\hat{\bm{P}_{\varepsilon}}$. 
\begin{equation}
\hat{\bm{P}_{\varepsilon}} = \text{DBSCAN}(\, \bm{P}, \, \varepsilon,  \, \text{MinPts} \, )
\end{equation}
where $\varepsilon$ is the maximum distance between two points, and  \text{MinPts} is minimum number of points required to form a dense region, which is set as 2 empirically.

\faHandPointRight[regular] \textbf{Step 4. Motion-Prior-Driven Pseudo Labels Filtering.}
For each trajectory $H_i$ intersecting the area $\hat{\bm{P}_{\varepsilon}}$, we assume that all RoIs $(B^{i}_{k}, C^{i}_{k}, S^{i}_{k})$ within $H_i$ should share the same classification label. To this end, we update the class of each RoI in $H_i$ with the most frequent class within this moving trajectory. More precisely, we calculate the frequency $f^i_c$ of each class $c$, and identify the most frequently occurring class as $C^i = \operatorname*{argmax}_c f^i_c$. We then replace the $C^i_k$ of all RoIs in this trajectory with the unified $C^i$, ensuring that all RoIs within the trajectory share the same class, denoted as $(B^{i}_{k}, C^{i}, S^{i}_{k}) \in H_i$.
This strategy leads to more accurate classification results for the pseudo labels.

%% file: section/experiment.tex
\section{Experimental Results} \label{sec:experiment}

\input{table/table_1}

In this section, we elaborate the experiments conducted to validate our proposed framework.

\subsection{Implementation, Dataset \& Evaluation Metrics}

\hspace*{4mm} \textbf{Implementation.}
Our proposed detection architecture is built on MMDetection~\cite{chen2019mmdetection} using the PyTorch~\cite{paszke2019pytorch} deep-learning framework. 
The detection backbone encoder is initialised with the "Xavier" initialisation method~\cite{glorot2010understanding} and consists of a ResNet-50~\cite{he2016deep} pretrained on ImageNet~\cite{deng2009imagenet}. The attention module's remaining parameters are randomly initialised using the "Normal" initialisation method.
During training, we employ the AdamW optimiser~\cite{loshchilov2017decoupled} with an initial learning rate of $2\times 10^{-4}$ and step decay.
All experiments and ablation studies are trained for 24 epochs, with a batch size of 1, taking approximately 6 hours of training time on an NVIDIA A100 GPU with 80GB RAM.

\textbf{Data Description.} 
We evaluate the effectiveness of our  method using the TrafficCAM~\cite{deng2022trafficcam} benchmark dataset.
TrafficCAM is a challenging traffic camera dataset designed for traffic flow surveillance. The dataset consists of 2,102 traffic videos captured from various traffic cameras in diverse scenes and weather conditions. 
The TrafficCAM dataset has in total 2,102 videos, and each video in the dataset is 3 seconds long and consists of 30 frames per video.
The spatial size of the video ranges from  $352 \times 288$ to $1920\times1080$.
While 78 videos are fully annotated for the entire duration, 2,024 videos are only annotated for the first frame.
For the fully-supervised setting, we use all the frames from the 27 annotated videos and the first frame of the remaining 2,024 videos for training. The remaining 51 annotated videos are used for testing. In the semi-supervised setting, we use the unlabelled 29 frames from the 2,024 videos for training along with the fully annotated 27 videos.

\textbf{Data Pre-Processing.}
We adopt the default setting in MMDetection and use the same data augmentation strategy as~\cite{ren2015faster} to increase the diversity of the dataset during training. Specifically, images are resized to $1333 \times 800$ and randomly flipped horizontally. During testing, we  resize the images to a unified size of $1333 \times 800$.

%\noindent
\textbf{Evaluation Metrics}. 
To evaluate the object detection performance, we follow the evaluation protocol used in object detection methods \cite{wu2019sequence} and employ six types of mean average precision (mAP). The six types of mAP are "mAP", "mAP @50", "mAP @75", "mAP small", "mAP medium", and "mAP large". mAP measures the average precision across multiple intersection over union (IoU) thresholds of object detection models. mAP@50 and mAP@75 are variants of mA that take IoU thresholds as 0.5 and 0.75, respectively. mAP small, mAP medium, and mAP large are variants of mAP that consider the object detection task's difficulty level based on the object size in the dataset. Higher mAP scores indicate better video object detection results. These additional metrics help identify the model's strength in detecting small objects, large objects, or both.

\subsection{Results \& Discussion}

\hspace{4mm}\textbf{Fully-Supervised Methods Comparison.} We start our evaluation by comparing our technique against existing SOTA video object detection methods, namely~DFF~\cite{zhu2017deep}, FGFA~\cite{zhu2017flow}, SELSA~\cite{wu2019sequence}, TRA~\cite{gong2021temporal}. 
Specifically, DFF~\cite{zhu2017deep} extracts deep features from frame pairs and warps them to a common reference frame using optical flow. FGFA~\cite{zhu2017flow} builds upon DFF and proposes a flow-guided feature aggregation module. SELSA~\cite{wu2019sequence} addresses temporal misalignment in VOD by introducing a latent sequential embedding module. TRA~\cite{gong2021temporal} captures object dependencies using a temporal relation module. We retrained these methods using unified training parameters and implemented the video object detection baselines using MMDetection~\cite{chen2019mmdetection}.

Table~\ref{table_1_numerical_comparision} summarises the mAP scores of our proposed technique and SOTA fully-supervised VOD methods on the TrafficCAM dataset. Our proposed video-level method achieves the best performance across all six mAP scores when compared with all existing SOTA methods, including image-level and video-level models. We observe that all proposed models perform better on detecting larger objects, where mAP$_l$ is the highest, followed by mAP$_m$, and the lowest score is mAP$_s$ for the same methods. In the upper half of the table, we notice that the best-performing image-level method on each metric is different. The two-staged method, Double Heads~\cite{wu2020rethinking}, performs the best on mAP score and mAP on medium-sized object detection. Faster RCNN~\cite{ren2015faster} and Guided Anchoring~\cite{wang2019region} achieve the best results on large objects and small objects, respectively, among all existing image object detection methods.

In video-level models, DFF~\cite{zhu2017deep} and FGFA~\cite{zhu2017flow} have significantly lower mAP scores than the other methods. These two methods perform substantially low on detecting small size objects, where both mAP$_s$ scores are under 0.5\%. This could be due to two reasons: firstly, the spatial information that DFF utilises is not guided with traffic-specific information; secondly, there is no attention mechanism implemented in these two methods. By adding attention mechanisms, the performance of SELSA~\cite{wu2019sequence} and TRA~\cite{gong2021temporal} that consider temporal relations between frames improves all six mAP scores by at least 5\%. Our proposed method that embeds with motion prior guided attention pushes the mAP scores even further, with scores of more than 1.5\%.

\input{table/table_2}

\begin{figure*}[t!]
    \centering
    \includegraphics[width=\textwidth]{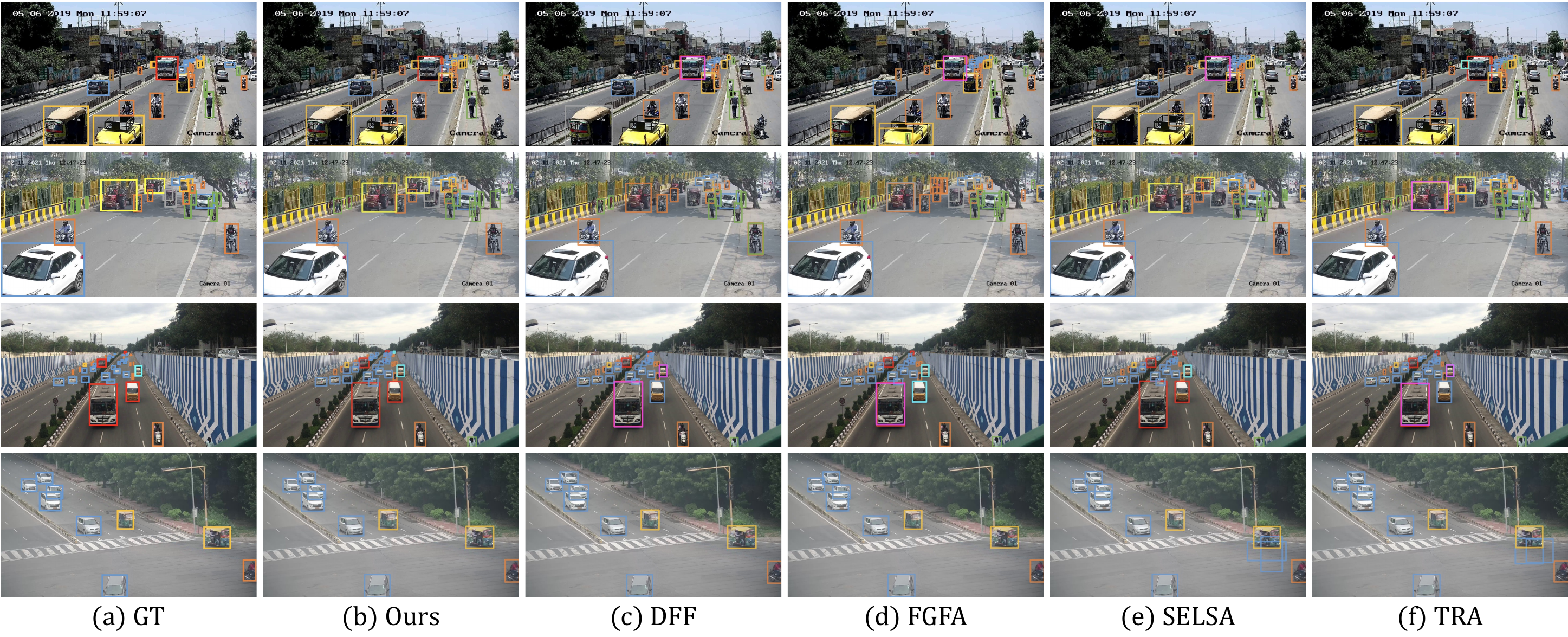}
    \vspace{-6mm}
    \caption{Visual comparison of traffic video object detection results obtained from our Motion Prior Attention method and other comparison methods. (c-f) represents the video models compared in Table~\ref{table_1_numerical_comparision}. Detailed video detection results can be found in the supplementary material. Different colors of bounding boxed denotes different classes. Detailed video detection results can be found in the supplementary material.
    }
    \label{fig:num_instance}
    \vspace{-3mm}
\end{figure*}

\textbf{Semi-Supervised Techniques Comparison.} In addition to fully-supervised settings, we also explored training our methods using our semi-supervised approach described in Section~\ref{subsec:semi-sup}, and compared with training our supervised methods on two SOTA semi-supervised methods for video object detection, STAC~\cite{sohn2020simple} and SoftTeacher~\cite{xu2021end}.

Table~\ref{table_2_numerical_comparision} presents a comparison of the performance of different semi-supervised frameworks using our proposed motion prior. The baseline refers to our proposed motion prior on a fully-supervised setting. We observed that in the STAC~\cite{sohn2020simple} framework, the additional 8,689 unlabelled frames improved the performance on all six metrics, but all scores were below 1\%. However, the SoftTeacher~\cite{xu2021end} method failed to surpass the baseline performance on mAP@50 and mAP$_l$, but obtained higher scores on the other 4 mAP measurements. Our proposed semi-supervised framework that embeds the motion prior further improved the SOTA semi-supervised methods, outperforming all compared methods in all six scores.

\textbf{Visual Performance Evaluation.}  To provide a more comprehensive evaluation of our proposed technique, we include a set of visual comparisons against existing methods in Figure 3. In a closer look at the results, we observe that our technique outperforms the compared methods. Specifically, all other techniques fail to correctly recognise certain classes, while our proposed technique demonstrates greater accuracy. For instance, in the first row, all techniques except for TRA fail to recognise several objects, such as the one enclosed in the red box, and TRA also fails in other cases, such as in the third row. Moreover, in the second column, the tractor is misidentified in all other techniques except ours. Additionally, our technique demonstrates greater prediction certainty compared to other methods. This is evident in other techniques, which tend to produce false positive bounding boxes in regions where objects are not present, as seen in the last row for techniques such as SELSA and TRA.

\input{table/table_4_ablation_study}

\begin{figure*}[t!]
    \centering
    \includegraphics[width=\textwidth]{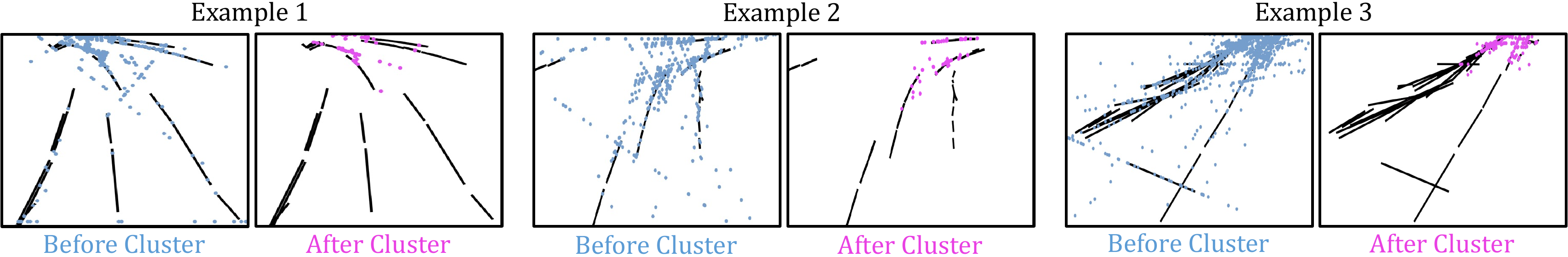}
    \caption{Visualisation of our pseudo label filtering strategy utilising motion prior. In each example, the left image displays the trajectories and intersection points (highlighted in blue), while the right image highlights the area centred around the vanishing point (highlighted in pink). After the clustering process, outlier intersection points are excluded. %, such as those appearing on the opposite side of the image in Example 1 or those originating from invalid trajectories in Examples 2 and 3.
    }
    \label{fig:clustering_results}
    \vspace{-3mm}
\end{figure*}

Overall, our technique is strongly supported by both numerical and visual evidence. Numerically, our results demonstrate superior performance compared to existing techniques, as indicated by higher mAP scores across various evaluation metrics. Visual comparisons also reveal the effectiveness of our approach, showcasing accurate object recognition and a higher level of certainty in predictions. These combined findings reinforce the robustness and efficacy of our proposed technique in addressing the challenges of the task at hand.

\textbf{Ablation Study on Motion Prior Attention.} We also conducted a series of ablation studies to validate the impact of our proposed techniques. We analyzed the efficacy of our proposed motion-prior attention by comparing it against several scenarios: employing no mask (default self-attention), a softmax operation, and a binary mask, respectively. 
Table~\ref{table_4_ablation_study_1} presents the performance of the self-attention module under these different settings. 
Using no mask or softmax shows minimal effect.
Notably, mask binarization improves performance but tends to aggressively remove other information, making it less versatile than our method.
In contrast, our motion-prior attention underscores the enhanced capability to more accurately capture trajectories in traffic videos and produce more effective features.

\input{table/table_3}

\textbf{Ablation Study on Motion-Prior-Enhanced Pseudo-Labeling.} For the motion-prior-enhanced pseudo-labeling, we investigated the impact of two important parameters: the selection of $\varepsilon$, and the number of frames $d$ used for constructing trajectories. Table~\ref{table_3_ablation_study} presents the results of our experiments with varying ranges of $\varepsilon$ and $d$.
Our findings indicate that the selection of $\varepsilon$ has a significant impact on the clustering results. If $\varepsilon$ is too small, the algorithm may identify too many small clusters as clusters, while a large value of $\varepsilon$ may merge multiple clusters into a single one or even consider all data points as a single cluster. We empirically found that a value of $1$ gives the best performance.
Regarding the value of $d$, we observed that in real-world scenarios, objects do not follow perfect trajectories, and a large value of $d$ may decrease the correction of the pseudo labels. On the other hand, a small value of $d$ may lead to a reduction in performance as several segments are generated.

To enhance the understanding of the pseudo-labelling filter, we present a set of visualisation results in Figure~\ref{fig:clustering_results}. We carefully selected three representative cases that demonstrate the accurate identification of vanishing point centred area. These visualisations provide a clear illustration of the effectiveness of our proposed filter in accurately identifying and distinguishing these critical points in the data.

\subsection{Limitation and Future Works}

The domain of autonomous driving frequently utilizes traffic videos to enhance auto-drive technology development. These videos often represent a driver's perspective, capturing the traffic scenario as experienced by a car, such as “in-car videos”. In contrast, our approach leverages traffic videos captured by stationary cameras positioned on roads, offering a more comprehensive, global view of traffic situations. Our methodology mainly targets this latter scenario, as the straight-line motion prior we investigated is more prominent in these types of traffic videos. Experiments conducted with the first type of video dataset, such as NuScenes~\cite{caesar2020nuscenes} and Waymo Open Dataset~\cite{sun2020scalability}, did not demonstrate a substantial performance improvement over SOTA methods, nor did it show a decline. This was anticipated, as in-car videos typically do not exhibit the straight-line motion prior that our study focuses on. Despite the prevalent use of in-car videos in autonomous driving, it is crucial to recognize the value of stationary camera traffic videos, as their global perspective can significantly augment autonomous driving systems.

Given these considerations, a promising future research direction is the development of a multimodal framework that efficiently integrates and utilizes both in-car and global view traffic videos to advance autonomous driving tasks. Furthermore, while our research emphasizes the commonly observed motion pattern in traffic videos, other scenarios display different motion patterns, such as those seen in starling flocks. The mathematical modeling of these complex motion patterns and integration into our proposed paradigm, offers a fascinating direction for future research.

%% file: table/table_1.tex
\begin{table*}[t!]
\centering
\resizebox{0.64\textwidth}{!}{
    \begin{tabular}{cc|cccccc}
        \hline \toprule
        \multicolumn{2}{c|}{\textsc{Methods}}                                                                                   & \multicolumn{6}{c}{\textsc{Evaluation Metrics}}                           \\ \hline
        \multicolumn{1}{c|}{Types}                        & \multicolumn{1}{c|}{Models}                                         & $\text{mAP}$      & $\text{mAP}_{50}$ & $\text{mAP}_{75}$  & $\text{mAP}_{s}$   & $\text{mAP}_{m}$   & $\text{mAP}_{l}$         \\ \midrule

        \multicolumn{1}{c|}{\multirow{7}{*}{Image}}       & \multicolumn{1}{c|}{Faster RCNN~\cite{ren2015faster}}               & 0.446    & 0.597    & 0.509     & 0.253     & 0.549     & 0.664           \\
        \multicolumn{1}{c|}{}                             & \multicolumn{1}{c|}{Libra~\cite{pang2019libra}}                     & 0.447    & 0.650    & 0.504     & 0.266     & 0.552     & 0.625           \\
        \multicolumn{1}{c|}{}                             & \multicolumn{1}{c|}{Guided Anchoring~\cite{wang2019region}}         & 0.465    & 0.694    & 0.516     & 0.310     & 0.538     & 0.631           \\
        \multicolumn{1}{c|}{}                             & \multicolumn{1}{c|}{Dynamic R-CNN~\cite{zhang2020dynamic}}          & 0.461    & 0.672    & 0.512     & 0.271     & 0.556     & 0.650           \\
        \multicolumn{1}{c|}{}                             & \multicolumn{1}{c|}{SABL~\cite{wang2020side}}                       & 0.471    & 0.648    & 0.531     & 0.280     & 0.568     & 0.660           \\
        \multicolumn{1}{c|}{}                             & \multicolumn{1}{c|}{Double Heads~\cite{wu2020rethinking}}           & 0.476    & 0.684    & 0.522     & 0.299     & 0.569     & 0.645           \\ \midrule[0.8pt]

        \multicolumn{1}{c|}{\multirow{5}{*}{Video}}       & \multicolumn{1}{c|}{DFF~\cite{zhu2017deep}}                         & 0.259    & 0.364    & 0.294     & 0.051     & 0.304     & 0.589           \\ 
        \multicolumn{1}{c|}{}                             & \multicolumn{1}{c|}{FGFA~\cite{zhu2017flow}}                        & 0.260    & 0.369    & 0.296     & 0.049     & 0.310     & 0.589           \\ 
        \multicolumn{1}{c|}{}                             & \multicolumn{1}{c|}{SELSA~\cite{wu2019sequence}}                    & 0.453    & 0.654    & 0.522     & 0.260     & 0.571     & 0.644           \\ 
        \multicolumn{1}{c|}{}                             & \multicolumn{1}{c|}{TRA~\cite{gong2021temporal}}                    & 0.464    & 0.677    & 0.513     & 0.274     & 0.552     & 0.653           \\ \cmidrule{2-8} 
        \multicolumn{1}{c|}{}                             & \multicolumn{1}{c|}{Ours}                                           & \textbf{0.496}   & \textbf{0.720}   & \textbf{0.566}   & \textbf{0.333}   & \textbf{0.591}  & \textbf{0.669}         \\ \bottomrule
        \end{tabular}
}
\vspace{-1mm}
\caption{Performance comparison between our Motion Prior Attention technique and state-of-the-art methods in a fully supervised setting on the TrafficCAM dataset. mAP denotes the mean Average Precision, where a higher score indicates superior performance. 
}
\label{table_1_numerical_comparision}
\vspace{-3mm}
\end{table*}
%}

%% file: table/table_2.tex
\begin{table}[t!]
\centering
\resizebox{\linewidth}{!}{
    \begin{tabular}{c|cccccc} \hline \toprule
        \multicolumn{1}{c|}{\multirow{2}{*}{\textsc{}}}                           & \multicolumn{6}{c}{\textsc{Evaluation Metrics}}                                                         \\ \cline{2-7}
                                                                                & $\text{mAP}$      & $\text{mAP}_{50}$ & $\text{mAP}_{75}$  & $\text{mAP}_{s}$   & $\text{mAP}_{m}$   & $\text{mAP}_{l}$                                       \\ \midrule
        \multicolumn{1}{l|}{Baseline}                                           & 0.496    & 0.720    & 0.566     & 0.333     & 0.591     & 0.669                                         \\ \midrule
        \multicolumn{1}{l|}{T1~\cite{sohn2020simple}}                        & 0.505    & 0.726    & 0.571     & 0.338     & 0.599     & 0.670                                         \\ 
        \multicolumn{1}{l|}{T2~\cite{xu2021end}}                      & 0.501    & 0.718    & 0.568     & 0.340     & 0.597     & 0.668                                         \\ \midrule
        \multicolumn{1}{l|}{Ours}                                              & \textbf{0.521} & \textbf{0.733} & \textbf{0.586} & \textbf{0.355} & \textbf{0.621} & \textbf{0.672}     \\ \bottomrule
        \end{tabular}
}
\vspace{-1mm}
\caption{Comparison of our Motion-Prior-Driven Pseudo-Labelling strategy against other methods in semi-supervised settings on the TrafficCAM dataset. mAP denotes the mean Average Precision, where a higher score indicates superior performance. The baseline represents the fully-supervised model derived from our methodology. We denote STAC technique as T1, and SoftTeacher as T2.
}
\label{table_2_numerical_comparision}
\vspace{-4mm}
\end{table}

%% file: table/table_4_ablation_study.tex
\begin{table}[t!]
\centering
\resizebox{\linewidth}{!}{
    \begin{tabular}{c|cccccc} \hline \toprule
        \multicolumn{1}{c|}{\multirow{2}{*}{\textsc{}}}                           & \multicolumn{6}{c}{\textsc{Evaluation Metrics}}                                                         \\ \cline{2-7}
                                                                                & $\text{mAP}$      & $\text{mAP}_{50}$ & $\text{mAP}_{75}$  & $\text{mAP}_{s}$   & $\text{mAP}_{m}$   & $\text{mAP}_{l}$                                       \\ \midrule
        \multicolumn{1}{l|}{No Mask}                    & 0.453  & 0.654   & 0.522   & 0.260   & 0.571   & 0.644                                         \\ 
        \multicolumn{1}{l|}{Softmax}    & 0.454  & 0.660   & 0.524   & 0.265   & 0.573   & 0.649                                         \\ 
        \multicolumn{1}{l|}{Binary}         & 0.477  & 0.679   & 0.518   & 0.271   & 0.578   & 0.655                                         \\ \midrule
        \multicolumn{1}{l|}{Ours}                        & \textbf{0.496}   & \textbf{0.720}    & \textbf{0.566}  & \textbf{0.333}  & \textbf{0.591} & \textbf{0.669}   \\ \bottomrule
        \end{tabular}
}
\vspace{-1mm}
\caption{Ablation study on different mask configurations in the motion-prior attention module.}
\label{table_4_ablation_study_1}
\vspace{-4mm}
\end{table}

%% file: table/table_3.tex
\begin{table}[t]
\centering
\vspace{0mm}
\resizebox{1\linewidth}{!}{
\begin{tabular}{cc|cccccc} \hline \toprule
    \multicolumn{2}{c|}{}                                                                   & \multicolumn{6}{c}{\textsc{Evaluation Metrics}}    \\ \hline
    \multicolumn{1}{{p{0.3cm}|}}{\,$d$}        & \multicolumn{1}{{p{0.3cm}|}}{\,\,$\varepsilon$}                  & $\text{mAP}$      & $\text{mAP}_{50}$ & $\text{mAP}_{75}$  & $\text{mAP}_{s}$   & $\text{mAP}_{m}$   & $\text{mAP}_{l}$ \\ \midrule
    \multicolumn{1}{c|}{3}                   & \multicolumn{1}{c|}{0.5}                               & 0.510   & 0.724    & 0.577     & 0.338     & 0.611    & 0.655   \\ 
    \multicolumn{1}{c|}{5}                   & \multicolumn{1}{c|}{0.5}                               & 0.517   & 0.730    & 0.576     & 0.349     & 0.611    & 0.662   \\
    \multicolumn{1}{c|}{8}                   & \multicolumn{1}{c|}{0.5}                               & 0.509   & 0.718    & 0.571     & 0.337     & 0.608    & 0.654   \\ \midrule
    \multicolumn{1}{c|}{3}                   & \multicolumn{1}{c|}{1}                                 & 0.512   & 0.721    & 0.575     & 0.346     & 0.608    & 0.666   \\ 
    \multicolumn{1}{c|}{\textbf{5}}              & \multicolumn{1}{c|}{\textbf{1}}                     & \textbf{0.521} & \textbf{0.733} & \textbf{0.586} & \textbf{0.355} & \textbf{0.621} & \textbf{0.672}  \\          
    \multicolumn{1}{c|}{8}                   & \multicolumn{1}{c|}{1}                                 & 0.510   & 0.712    & 0.571     & 0.340     & 0.603    & 0.665   \\ \midrule
    \multicolumn{1}{c|}{3}                   & \multicolumn{1}{c|}{1.5}                               & 0.492   & 0.711    & 0.560     & 0.324     & 0.578    & 0.653   \\ 
    \multicolumn{1}{c|}{5}                   & \multicolumn{1}{c|}{1.5}                               & 0.497   & 0.718    & 0.566     & 0.330     & 0.589    & 0.660   \\ 
    \multicolumn{1}{c|}{8}                   & \multicolumn{1}{c|}{1.5}                               & 0.490   & 0.710    & 0.554     & 0.321     & 0.579    & 0.655   \\ \bottomrule
\end{tabular}
}
\vspace{-1mm}
\caption{Ablation study for different components within our proposed pseudo-label filtering strategy.}
\label{table_3_ablation_study}
\vspace{-3mm}
\end{table}

%% file: section/conclusion.tex
\section{Conclusion} \label{sec:conclusion}
In conclusion, we have presented a pioneering approach to traffic video object detection that harnesses the unique features of traffic videos. Our method takes advantage of the fixed camera positions in traffic videos, which provides a strong motion prior, indicating that objects will move in a specific direction. We have introduced two innovative techniques that utilise this motion prior to enhance the performance of both fully-supervised and semi-supervised traffic video object detection. The first technique involves the creation of a self-attention module that establishes robust temporal correlations between frames in a fully-supervised setting. The second technique focuses on designing a pseudo-labelling mechanism to eliminate noisy pseudo labels within a semi-supervised context. Both methods, grounded in the motion prior, outshine existing approaches that disregard the task-specific knowledge inherent in traffic videos. Our comprehensive evaluation demonstrates that our framework consistently surpasses current state-of-the-art methods.

\section*{Acknowledgement} 

LL gratefully acknowledges the financial support from a GSK Ph.D. Scholarship and a Girton College Graduate Research Fellowship at the University of Cambridge, and the support from Oracle Ph.D. Project Award.
AIAR acknowledges support from CMIH (EP/T017961/1) and CCIMI, University of Cambridge. This work was supported in part by Oracle Cloud credits and related resources provided by Oracle for Research.
CBS acknowledges support from the Philip Leverhulme Prize, the Royal Society Wolfson Fellowship, the EPSRC advanced career fellowship EP/V029428/1, EPSRC grants EP/S026045/1 and EP/T003553/1, EP/N014588/1, EP/T017961/1, the Wellcome Innovator Awards 215733/Z/19/Z and 221633/Z/20/Z, the European Union Horizon 2020 research and innovation programme under the Marie Skodowska-Curie grant agreement No. 777826 NoMADS, the Cantab Capital Institute for the Mathematics of Information and the Alan Turing Institute.